\documentclass{INTERSPEECH2023}
\usepackage{multirow}

\interspeechcameraready

\title{Automatic Heteronym Resolution Pipeline Using RAD-TTS Aligners}
\name{Jocelyn Huang$^1$\sthanks{*Equal contribution.}, Evelina Bakhturina$^1$$^*$, Oktai Tatanov$^2$\sthanks{$^\dagger$ Work done while at NVIDIA.}}
\address{$^1$NVIDIA, $^2$AXB Research}
\email{jocelynh@nvidia.com, ebakhturina@nvidia.com}

\begin{document}

\maketitle
 
\begin{abstract}
Grapheme-to-phoneme (G2P) transduction is part of the standard text-to-speech (TTS) pipeline.
However, G2P conversion is difficult for languages that contain heteronyms -- words that have one spelling but can be pronounced in multiple ways.
G2P datasets with annotated heteronyms are limited in size and expensive to create, as human labeling remains the primary method for heteronym disambiguation. We propose a RAD-TTS Aligner-based pipeline to automatically disambiguate heteronyms in datasets that contain both audio with text transcripts. The best pronunciation can be chosen by generating all possible candidates for each heteronym and scoring them with an Aligner model. The resulting labels can be used to create training datasets for use in both multi-stage and end-to-end G2P systems. 
\end{abstract}

\section{Introduction}
\label{sec:intro}

Modern text-to-speech (TTS) models can learn pronunciations from raw text input and its corresponding audio data, but in languages such as English, phonemes provide more precise pronunciation information than graphemes. As a result, many TTS systems use phonemic input during training to directly access and correct pronunciations for new vocabulary at inference time. One of the hardest problems for grapheme-to-phoneme (G2P) systems is the resolution of heteronyms, i.e., words that have a single spelling but different pronunciations. For example, \textit{``read"} in \textit{“I will read the book”} vs. \textit{“She read her project last week”}. Some heteronyms, such as \textit{``bass"}, have multiple pronunciations with the same part of speech, and they need to be disambiguated based on semantic context.

In this work, we focus on the heteronym disambiguation task and propose a pipeline for labeling heteronyms in training data for both multi-stage and end-to-end (E2E) G2P models.
Some multi-stage G2P systems \cite{g2pE2019, espeakng} use a set of rules for heteronym disambiguation, but high-quality rule-based systems require expert knowledge and are difficult to scale and maintain. An alternative machine learning approach for heteronym disambiguation is to treat this task as a part-of-speech tagging or a classification problem \cite{yarowsky1997homograph, gorman-etal-2018-improving}. Emerging E2E G2P systems use sentence-level training data \cite{vrezavckova2021t5g2p, ploujnikov2022soundchoice} and aim to handle out-of-vocabulary (OOV) and heteronyms in a single pass. Neural multi-stage and E2E solutions for heteronym disambiguation require labeled data where heteronyms appear in context, but unfortunately, there is a dearth of such data.

Due to the domain expertise required for labeling phonemes, G2P datasets are few and far between. In datasets like TIMIT \cite{timit} and The Buckeye Speech Corpus \cite{buckeye}, phoneme transcriptions of audio are provided along with grapheme transcriptions. In TIMIT, transcriptions were human-verified, but the number of unique sentences is too small to train a G2P model. The Buckeye Speech Corpus consists of around 26 hours of conversational speech that was transcribed and phonemically labeled. Since the phoneme labels were automatically generated from the audio, the labels are noisy and sometimes contain alignment errors despite some corrections made by human research assistants, which makes the dataset more unreliable for G2P training.

To our knowledge, the Wikipedia Homograph Data \cite{gorman-etal-2018-improving} (WikiHomograph) is the only open-source dataset with a sufficient number of samples to train a neural model for heteronym disambiguation. WikiHomograph is a text-only dataset where each sample is an entire sentence with a labeled heteronym. Unfortunately, this dataset does not contain a comprehensive list of English homographs. Moreover, some pronunciations in the WikiHomograph set of heteronyms are significantly underrepresented, leading to class imbalance \cite{nicolis11homograph}. For example, the corpus contains multiple sentences with the noun form of the heteronyms ``desert", ``addict" and ``subject" and no samples with the verb forms. The WikiHomograph dataset was annotated by linguists, and manual annotation remains the mainstream method of data creation.
In addition, some preprocessing is required to train an E2E G2P model on the WikiHomograph dataset, as only the target homograph is labeled in each example sentence. \cite{ploujnikov2022soundchoice} uses CMUdict \cite{cmudict} to label known words while dropping sentences with OOV words.

As a heteronym data augmentation technique, Nishiyama et al. \cite{nishiyama-etal-2018-dataset} introduced a method to match each sense of a heteronym to a synonymous word with a unique pronunciation and to substitute the heteronym for its synonym in a text corpus. This method requires a large textual database for queries, as well as expert knowledge and evaluators to confirm that the resulting sentences are correct. As the method was applied to Japanese heteronyms, there is no available data for English.
Other relevant methods of heteronym resolution and verification include the morphological rewriting rules \cite{matsuoka} and the context-dependent phone-based HMMs that use acoustic features \cite{lu2008}. \cite{tatanov2022mixer} skips the phoneme representation in lieu of passing graphemes into a language model to generate its text representation. We plan to add these to our paper to address this broader context.

\begin{figure*}[h]
    \includegraphics[width=\textwidth]{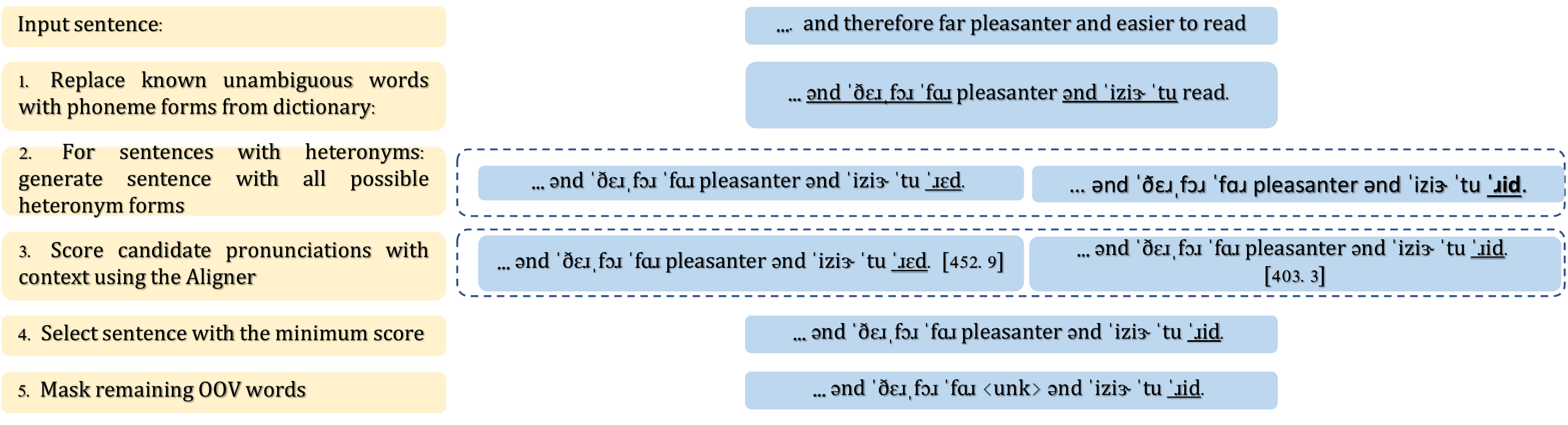}
    \caption{Data labeling pipeline for sentence-level G2P model training includes the following steps: 1) Input text. 2) Replace known unambiguous words with phoneme forms from the dictionary. 3) For sentences with heteronyms: generate sentences with all possible heteronym forms. 4) Score candidate pronunciations with context using the Aligner. 5) Select a sentence with the minimum score. 6) Mask remaining OOV words.
}
    \label{fig:pipeline}
\end{figure*}

\begin{figure*}[t]
    \includegraphics[width=\textwidth]{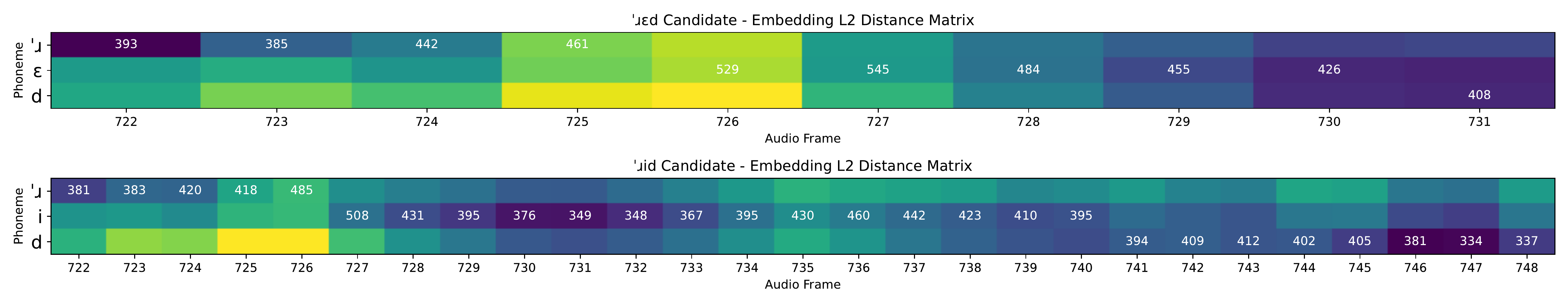}
    \caption{A comparison between the L2 distance matrices between the aligned text and audio embeddings when disambiguating the word \textit{``read"} from the entry: \textit{``... and therefore far pleasanter and easier to read"}. Values shown correspond to the audio frames that were aligned with each text token, and the average distance is taken across this diagonal to find the overall score for a given pronunciation; the rest of the values are disregarded. The average embedding distances for \textit{/\textipa{\*rEd}/} and \textit{/\textipa{\*rid}/} are 452.9 and 403.3, respectively. The latter one would be picked, as it is closer to the audio embedding across the aligned frames.}
    \label{fig:emb_distance}
\end{figure*}


We propose an automatic heteronym disambiguation approach that can generate examples for underrepresented or missing heteronym forms.
Our proposed pipeline annotates speech data with heteronym phoneme labels automatically. The labeled sentences can then be used in conjunction with dictionary lookups for unambiguous known words and ``$<$unk$>$" tokens for OOV words to create training data for neural G2P or heteronym classification models without human labeling. To get target phonemic labels for heteronyms, we train the RAD-TTS Aligner \cite{radtts_aligner} on transcribed audio data. Then we use the Aligner to score possible heteronym pronunciation options and choose the one that matches the corresponding audio best.
To evaluate the quality of generated data, we train a BERT-based classification model and E2E ByT5 G2P model. The results show that the proposed data augmentation technique improves heteronym disambiguation accuracy for both models. We release code\footnote{\url{https://github.com/NVIDIA/NeMo}} 
and all aligner-generated and hand-checked data for heteronym disambiguation model training.


\section{Heteronym resolution pipeline}

We propose using a RAD-TTS Aligner \cite{radtts_aligner} model to automatically select correct heteronym forms. The RAD-TTS Aligner \cite{radtts_aligner} is a speech-to-text alignment model based on the alignment mechanism introduced in RAD-TTS \cite{shih2021rad}, which allows for easy visualization and human-understandable scores when comparing candidate pronunciations. The Aligner takes a mix of graphemes and phonemes as input: phonemes for known unambiguous words and graphemes for ambiguous or OOV words. It learns to align text tokens and audio frame encodings using the $L_2$ distance between the representations, generating a soft alignment that can be converted to a hard alignment using the Viterbi algorithm. 

These hard alignments between text tokens and audio frames can be used in tandem with the predicted $L_2$ distance matrix in order to determine the distances between a token encoding and each of its corresponding audio frames' encodings. Thus, given a word $T$ consisting of $N$ input tokens $t_1, ..., t_N$, where token $t_i$ has been aligned with $M_i$ audio frames $a_1^{(i)}, ..., a_{M_i}^{(i)}$ out of audio $A$, the average distance, $D_{avg}$, between a word and the audio can be found as:
\begin{equation}
D_{avg}\big(T, A\big) = \frac{\sum\limits_{i=1}^N \sum\limits_{j=1}^{M_i} L_2(enc\_t_i, enc\_a_{j}^{(i)})}{\sum\limits_{i=1}^N M_i}
\end{equation}
In essence, the average distance between a word and its acoustic form is a sum of distances between its constituent tokens and their aligned audio frames, divided by the number of audio frames corresponding to the word. We can use these distances to disambiguate heteronyms with an audio sample.

Figure \ref{fig:pipeline} shows the proposed automatic phoneme-labeling process for generating sentences with disambiguated heteronyms for sentence-level G2P model training. We first convert known unambiguous words to their phonetic pronunciations with dictionary lookups. This work uses the CMUdict training split defined in \cite{zhu2022byt5}. OOV words are left as graphemes. Next, we generate multiple candidates by substituting the heteronym with each possible phonemic form in the dictionary. Then, we pass each candidate along with the corresponding audio file through a trained Aligner model to automatically label heteronyms by picking the pronunciation whose phoneme encodings are closer on average to the audio encodings, i.e., smallest $D_{avg}$. Figure \ref{fig:emb_distance} shows an example of the alignments and distances for two potential pronunciations of \textit{``read"} from an entry that ends \textit{``and therefore far pleasanter and easier to read."} Using this method, we can disambiguate all known heteronyms in our speech dataset. Finally, we mask out OOV words with a special masking token, ``$<$unk$>$", and force the G2P model to produce the same masking token as a phonetic representation during training. During inference, the model generates phoneme predictions for OOV words without emitting the masking token as long as this token is not included in the grapheme input.

To control the quality of the disambiguated data, we propose thresholding with a confidence score that represents how much closer the best candidate pronunciation is to the audio. Specifically, the score is a normalized difference between the chosen candidate's L2 distance versus the least likely candidate's L2 distance. The confidence score of disambiguation is found by taking the difference between the highest and lowest L2 distances over all the candidates, then dividing it by the average between the highest and lowest L2 distances. For the example in Figure \ref{fig:emb_distance}, this would be 
$(452.9-403.3)/(452.9+403.3)/2) = 0.116$. 
The higher the score, the more likely it is for the disambiguation to be correct. We can now remove any samples with disambiguations that have confidence scores lower than the desired threshold. Once heteronym disambiguations have been performed, the sentences can then be converted to phonemes for use in sentence-level G2P training. As before, we use a dictionary lookup for known unambiguous words, and now we can replace heteronyms with the disambiguated phoneme form. Samples with OOV words can either be dropped, or OOV labels can be replaced with an $\langle$unk$\rangle$ token for training.

\section{Aligner training and dataset generation}
\label{heteronym_experiments}
We use the LJSpeech \cite{ljspeech17} and Hi-Fi TTS \cite{bakhturina21_interspeech} (speakers 9017 and 12787) datasets to generate G2P data with disambiguated heteronyms, and train one Aligner model per speaker.
Speaker 9017's data contains 57.8 hours and its Aligner model was trained for 250 epochs, speaker 12787 contains 27.1 hours and its Aligner model was trained for 400 epochs, and the LJSpeech model was trained for 1000 epochs on 22.8 hours of data. All models were trained on a single RTX 8000 GPU using the Adam optimizer, a learning rate of 0.001, and a weight decay of 1e-6. A Cosine Annealing scheduler was used, with a minimum learning rate of 5e-5 and a warmup ratio of 0.35.

For disambiguation, sentences without heteronyms were discarded. Aligner-disambiguated training sets of speakers 9017, 12787, and LJSpeech were compiled into the \textbf{Aug} set. We also created subsets of the data by filtering out samples where the Aligner confidence score was below a threshold value: \textbf{Aug-0.01\%} consists of samples with a confidence score of at least 0.01\%; similarly for thresholds of 0.02\% and 0.03\%. For each augmented subset, we created a ''balanced" version that aims to equalize the number of occurrences of each heteronym form in the combined WikiHomograph and Aug. training data to mitigate model bias (Table \ref{tab:aligner_stats}).

\begin{table}[]
\centering
\caption{Number of aligner-generated samples added depending on the confidence threshold values and balancing strategy.}
\resizebox{\columnwidth}{!}{
\begin{tabular}{l|c|c|c|c}
\hline
\multicolumn{1}{c|}{\textbf{Threshold}} & \textbf{0.00\%} & \textbf{0.01\%} & \textbf{0.02\%} & \textbf{0.03\%} \\ \hline
Num samples (bal)                             & 1230          & 794           & 620           & 572           \\ \hline
Num samples (non bal)                         & 3883          & 2939          & 2286          & 1805          \\ \hline
\end{tabular}}
\label{tab:aligner_stats}
\end{table}

\section{Evaluation}

\begin{table}[t]
\centering
\caption{True positives and false positives of each pronunciation of ``subject" as predicted by the speaker 9017 Aligner with various confidence thresholds.}
\resizebox{\columnwidth}{!}{%
\begin{tabular}{l|c|c|c|c|c}
\hline
\multirow{2}{*}{\textbf{``Subject" Eval}} & \multicolumn{2}{|c|}{\textbf{/\textipa{s@b"dZEkt}/ (v.)}}                      & \multicolumn{2}{c|}{\textbf{/\textipa{"s@bdZIkt}/ (adj./n.)}} & \multirow{2}{*}{\textbf{Total}}                  \\ \cline{2-5} 
& TP & FP & TP  & FP & \\ \hline
Threshold: 0.00\% & 1 & 30 & 48 & 0 & 79 \\ \hline
Threshold: 0.01\% & 1 & 5 & 25 & 0 & 31 \\ \hline
Threshold: 0.02\% & 1 & 1 & 13 & 0 & 15 \\ \hline
Threshold: 0.03\% & 0 & 0 & 4 & 0 & 4 \\ \hline
\end{tabular}
}
\label{tab:subject_disamb}
\end{table}

\begin{table}[t]
\centering
\caption{Accuracy on the heteronym disambiguation task of the BERT-based heteronym classification model on WikiHomograph and Hard evaluation sets depending on the amount and quality of the Aligner-generated augmented data.}
\begin{tabular}{lccc}
\hline
\multicolumn{1}{l|}{\multirow{2}{*}{\textbf{Training data}}} & \multicolumn{1}{c|}{\multirow{2}{*}{\textbf{Threshold}}} & \multicolumn{2}{c}{\textbf{Accuracy, \%}}                                \\ \cline{3-4} 
\multicolumn{1}{l|}{}                               & \multicolumn{1}{c|}{}                      & \multicolumn{1}{c|}{\textbf{WikiH}} & \multicolumn{1}{c}{\textbf{Hard}} \\ \hline
\multicolumn{1}{l}{WikiHomograph}                          & \multicolumn{1}{c}{-}                     & \multicolumn{1}{c}{98.70}      & \multicolumn{1}{c}{86.64}     \\ \hline
                                                     & 0.00\%                                       & 98.99                           & 89.63                          \\
+ Aligner data                                       & 0.01\%                                       & 98.97                           & 90.88                          \\
(no balance)                                         & 0.02\%                                       & \textbf{99.07}                           & \textbf{91.04}                          \\
                                                     & 0.03\%                                       & 98.97                           & 90.09                          \\ \hline
                                                     & 0.00\%                                       & 98.97                           & 83.02                          \\
+ Aligner data                                       & 0.01\%                                       & 99.05                           & 89.47                          \\
(balance)                                            & 0.02\%                                       & 99.03                           & 89.00                          \\
                                                     & 0.03\%                                       & 99.03                           & 89.46        \\  \bottomrule  
\end{tabular}
\label{tab:bert-cl}
\end{table}

\begin{table}[]
\centering
\caption{Evaluation of ByT5 E2E G2P model on heteronym disambiguation task (accuracy on WikiHomograph and Hard set) and on OOV (PER on CMUdict test split) depending on the Aligner-augmented data.}
\begin{tabular}{lcccc}
\hline
\multicolumn{1}{l|}{\multirow{2}{*}{\textbf{Training data}}} & \multicolumn{1}{c|}{{\textbf{Thres-}}} & \multicolumn{2}{c|}{\textbf{Accuracy, \%}}                         & \multicolumn{1}{c}{\textbf{CMUdict}} \\ \cline{3-5} 
\multicolumn{1}{l|}{}                               & \multicolumn{1}{c|}{\textbf{hold}}                      & \multicolumn{1}{c|}{\textbf{WikiH}} & \multicolumn{1}{c}{\textbf{Hard}} & \multicolumn{1}{|c}{\textbf{PER, \%}} \\ \hline
WikiH + CMU                                          & -                                          & 95.42                         & 79.72                     & 8.62                         \\ \hline
                                                     & 0.00\%                                       & 95.42                         & 83.02                     & 8.24                         \\
+ Aligner                                            & 0.01\%                                       & \textbf{96.10}                         & \textbf{85.85}                     & 8.97                         \\
(not balanced)                                       & 0.02\%                                       & 95.79                         & 82.08                     & 8.47                         \\
                                                     & 0.03\%                                       & 95.79                         & 83.02                     & \textbf{8.06}     \\ \bottomrule                   
\end{tabular}
\label{tab:byt5}
\end{table}

To assess the quality of heteronym resolution with the Aligner model, we hand-checked sentences from LJSpeech dev set, which contains 26 heteronyms. The LJSpeech Aligner model chose the grammatically correct candidate 23 times. However, two of the grammatically incorrect selections accurately reflected the pronunciation of the speaker. We also performed limited human evaluation of the heteronym labels derived from the Hi-Fi TTS speaker 9017 model for textit{``read"} and \textit{``subject"}. Out of 179 occurrences of the word \textit{``read,"} (87 /\textipa{\*rid}/, 92 /\textipa{\*rEd}/), the Aligner model picked the correct form 176 times (an accuracy of 98.3\%), with only three errors. However, it performs poorly on heteronym \textit{``subject"}, which has two forms that sound similar: /\textipa{s@b"dZEkt}/ and /\textipa{"s@bdZIkt}/. This can be mitigated by confidence thresholding, as seen in Table \ref{tab:subject_disamb}. We conclude that the Aligner model is highly dependent on the enunciation and pronunciation of the speaker, and is prone to error if the audio is noisy or if the speaker mispronounces a heteronym. It also tends to have trouble with heteronyms that have forms that sound similar, but this can be mitigated by confidence thresholding. 


We also manually verified heteronyms from the dev and test sets of the selected Hi-Fi TTS speakers. We then combined these samples with some proprietary sentences to create a test set that covers most of the heteronym forms missing from the evaluation set of the WikiHomograph dataset. This dataset (hereafter \textbf{Hard-eval}) contains 195 sentences and is used to evaluate the effect of the Aug data on the G2P models' performance.

To perform automatic quality estimation, we train a token classification BERT-based \cite{devlin2018bert} heteronym disambiguation model on the WikiHomograph dataset. The model takes a sentence as input, and then for every word, it selects a heteronym option out of the available dictionary forms. The model handles multiple heteronyms simultaneously. We mask irrelevant forms to disregard the model’s predictions for non-ambiguous words. E.g., given the input ``The Poems are simple to read and easy to comprehend.” the model scores possible {`read present` and `read past`} options for the word ``read”. We finetuned our model from pre-trained ``bert-base-cased"\footnote{{https://huggingface.co/bert-base-cased}} model for ten epochs on a 16GB GPU with batch size 32, the AdamW optimizer, a learning rate of 5e-5, a WarmupAnnealing scheduler, and a weight decay of 0.01.

Table \ref{tab:bert-cl} summarizes experiments with the BERT classification model trained on WikiHomograph data and various amounts of Aligner-generated data. The results are the averages of 3 runs. The highest accuracy was achieved on WikiHomograph and Hard-eval sets with ``non-balanced 0.02'' aligner data augmentation, 99.07\% and 91.04\%, respectively. Performance on the balanced set is more consistent on the WikiHomograph set (99+\%) and slightly below the best result. Non-balanced data augmentation leads to better results in the Hard-eval set than the performance with balanced data augmentation, 90+\% vs. about 89\%. We hypothesize that this is because the augmented data provides more non-Wikipedia examples with a vocabulary closer to the Hard-eval set. A confidence threshold of at least 0.01\% is recommended as it provides a higher quality of the augmented data; see the performance drop from 86.64\% to 83.02\% if no thresholding is used. The heteronym disambiguation task has a low tolerance towards errors as these errors propagate down to the text-to-speech pipeline. Using higher Aligner threshold values reduces the number of the augmented samples but assures a higher quality of the training data.

To check the validity of our sentence-level labeling pipeline on E2E G2P models, we follow \cite{vrezavckova2021t5g2p} and \cite{zhu2022byt5} and train a sentence-level ByT5 model G2P model.
The training data for our E2E G2P model consists of CMUdict \cite{cmudict} and WikiHomograph with various amounts of Aligner augmented data. We used the same CMUdict split proposed in \cite{zhu2022byt5} for labeling known words and ``$<$unk$>$" token for OOV words. We finetuned our model from pre-trained ``google/byt5-small"\footnote{{https://huggingface.co/google/byt5-small}} model for five epochs on eight 16GB GPUs with batch size 8, the AdamW optimizer, a learning rate of 1e-3, a WarmupAnnealing scheduler, and a weight decay of 0.01. Experiments with E2E ByT5 model (Table \ref{tab:byt5}) second the positive effect of the data augmentation while keeping the phoneme error rate (PER) on CMUDict test nearly the same. PER measures the generation capabilities of E2E G2P models.

\section{Conclusions}

In this paper, we propose a data augmentation method that can automatically disambiguate heteronyms to generate data for sentence-level G2P model training. This data labeling technique can be used to balance out existing heteronym forms in gold standard data, add new heteronyms without manual labeling, or simply create more training data as labeled heteronym data is scarce. The proposed method is also controllable using confidence threshold filtering, depending on whether a particular application may need more data with potentially lower quality, or high confidence labels at the cost of the number of samples generated. Additionally, we introduce a masking token that opens the door to sentence-level G2P model training without human annotation. 
We show through human evaluation and experimentation that the resulting automatically-generated data improves the performance of both BERT classification and E2E G2P systems. We hope that this method will help to remedy this lack of data both for more robust training and for more informative evaluation.

\bibliographystyle{IEEEtran}
\bibliography{mybib}

\begin{thebibliography}{10}
\providecommand{\url}[1]{#1}
\csname url@samestyle\endcsname
\providecommand{\newblock}{\relax}
\providecommand{\bibinfo}[2]{#2}
\providecommand{\BIBentrySTDinterwordspacing}{\spaceskip=0pt\relax}
\providecommand{\BIBentryALTinterwordstretchfactor}{4}
\providecommand{\BIBentryALTinterwordspacing}{\spaceskip=\fontdimen2\font plus
\BIBentryALTinterwordstretchfactor\fontdimen3\font minus
  \fontdimen4\font\relax}
\providecommand{\BIBforeignlanguage}[2]{{%
\expandafter\ifx\csname l@#1\endcsname\relax
\typeout{** WARNING: IEEEtran.bst: No hyphenation pattern has been}%
\typeout{** loaded for the language `#1'. Using the pattern for}%
\typeout{** the default language instead.}%
\else
\language=\csname l@#1\endcsname
\fi
#2}}
\providecommand{\BIBdecl}{\relax}
\BIBdecl

\bibitem{g2pE2019}
K.~Park and J.~Kim, ``{g2pE},'' \url{https://github.com/Kyubyong/g2p}, 2019.

\bibitem{espeakng}
R.~H. Dunn, ``{eSpeak NG},'' \url{https://github.com/espeak-ng/espeak-ng},
  2007.

\bibitem{yarowsky1997homograph}
D.~Yarowsky, ``Homograph disambiguation in text-to-speech synthesis,'' in
  \emph{Progress in speech synthesis}.\hskip 1em plus 0.5em minus 0.4em\relax
  Springer, 1997, pp. 157--172.

\bibitem{gorman-etal-2018-improving}
K.~Gorman, G.~Mazovetskiy, and V.~Nikolaev, ``Improving homograph
  disambiguation with supervised machine learning,'' in \emph{{LREC}}, 2018.

\bibitem{vrezavckova2021t5g2p}
M.~Řezáčková, J.~Švec, and D.~Tihelka, ``{T5G2P: Using Text-to-Text
  Transfer Transformer for Grapheme-to-Phoneme Conversion},'' in
  \emph{Interspeech}, 2021.

\bibitem{ploujnikov2022soundchoice}
A.~Ploujnikov and M.~Ravanelli, ``{SoundChoice: Grapheme-to-Phoneme Models with
  Semantic Disambiguation},'' in \emph{Interspeech}, 2022.

\bibitem{timit}
J.~S. Garofolo, L.~F. Lamel, W.~M. Fisher, J.~G. Fiscus, D.~S. Pallett, and
  N.~L. Dahlgren, ``{DARPA TIMIT} acoustic phonetic continuous speech corpus
  cdrom,'' 1993.

\bibitem{buckeye}
\BIBentryALTinterwordspacing
E.~Fosler-Lussier, L.~Dilley, N.~R. Tyson, and M.~A. Pitt, ``The buckeye corpus
  of speech: updates and enhancements.'' in \emph{INTERSPEECH}, 2007. [Online].
  Available:
  \url{http://dblp.uni-trier.de/db/conf/interspeech/interspeech2007.html#Fosler-LussierDTP07}
\BIBentrySTDinterwordspacing

\bibitem{nicolis11homograph}
M.~Nicolis and V.~Klimkov, ``Homograph disambiguation with contextual word
  embeddings for {TTS} systems,'' in \emph{Speech Synthesis Workshop (SSW 11)},
  2021.

\bibitem{cmudict}
``{The CMU Pronouncing Dictionary},''
  \url{http://www.speech.cs.cmu.edu/cgi-bin/cmudict/}, 2017.

\bibitem{nishiyama-etal-2018-dataset}
K.~Nishiyama, K.~Yamamoto, and H.~Nakajima, ``Dataset construction method for
  word reading disambiguation,'' in \emph{Pacific Asia Conference on Language,
  Information and Computation}, 2018.

\bibitem{matsuoka}
K.~Matsuoka, E.~Takeishi, H.~Asano, R.~Ichii, and Y.~Ooyama, ``Natural language
  processing in a japanese text-to-speech system for written-style texts,'' in
  \emph{Proceedings of IVTTA '96. Workshop on Interactive Voice Technology for
  Telecommunications Applications}, 1996, pp. 33--36.

\bibitem{lu2008}
H.~Lu, Z.~Ling, S.~Wei, Y.~Hu, L.~Dai, and R.~Wang, ``Heteronym verification
  for mandarin speech synthesis,'' in \emph{2008 6th International Symposium on
  Chinese Spoken Language Processing}, 2008, pp. 1--4.

\bibitem{tatanov2022mixer}
O.~Tatanov, S.~Beliaev, and B.~Ginsburg, ``Mixer-tts: non-autoregressive, fast
  and compact text-to-speech model conditioned on language model embeddings,''
  in \emph{ICASSP 2022-2022 IEEE International Conference on Acoustics, Speech
  and Signal Processing (ICASSP)}.\hskip 1em plus 0.5em minus 0.4em\relax IEEE,
  2022, pp. 7482--7486.

\bibitem{radtts_aligner}
R.~Badlani, A.~{Łańcucki}, K.~J. Shih, R.~Valle, W.~Ping, and B.~Catanzaro,
  ``One {TTS} alignment to rule them all,'' in \emph{ICASSP}, 2022.

\bibitem{shih2021rad}
K.~J. Shih, R.~Valle, R.~Badlani, A.~Lancucki, W.~Ping, and B.~Catanzaro,
  ``Rad-tts: Parallel flow-based {TTS} with robust alignment learning and
  diverse synthesis,'' in \emph{{ICML} Workshop on Invertible Neural Networks,
  Normalizing Flows, and Explicit Likelihood Models}, 2021.

\bibitem{zhu2022byt5}
J.~Zhu, C.~Zhang, and D.~Jurgens, ``{ByT5 model for massively multilingual
  grapheme-to-phoneme conversion},'' in \emph{Interspeech}, 2022.

\bibitem{ljspeech17}
K.~Ito and L.~Johnson, ``{The LJ Speech Dataset},''
  \url{https://keithito.com/LJ-Speech-Dataset/}, 2017.

\bibitem{bakhturina21_interspeech}
E.~Bakhturina, V.~Lavrukhin, B.~Ginsburg, and Y.~Zhang, ``{Hi-Fi Multi-Speaker
  English TTS Dataset},'' in \emph{Interspeech}, 2021.

\bibitem{devlin2018bert}
J.~Devlin, M.-W. Chang, K.~Lee, and K.~Toutanova, ``{BERT}: Pre-training of
  deep bidirectional transformers for language understanding,'' in
  \emph{"Conference of the North {A}merican Chapter of the Association for
  Computational Linguistics: Human Language Technologies"}, 2019.

\end{thebibliography}

\end{document}